\documentclass[unnumsec,webpdf,contemporary,large]{oup-authoring-template}%
\graphicspath{{Fig/}}

\theoremstyle{thmstyleone}%
\theoremstyle{thmstyletwo}%
\theoremstyle{thmstylethree}%

\begin{document}

\journaltitle{Journal Title Here}
\DOI{DOI HERE}
\copyrightyear{2019}
\pubyear{2019}
\access{Advance Access Publication Date: Day Month Year}
\appnotes{Paper}

\firstpage{1}

%\subtitle{Subject Section}

\title[GNN for drug combination screening]{DeepDDS: deep graph neural network with attention mechanism to predict synergistic drug combinations}

\author[1,\dag]{Jinxian Wang}
\author[2,\dag]{Xuejun Liu}
\author[1]{Siyuan Shen}
\author[1,3,$\ast$]{Lei Deng}
\author[2,$\ast$]{Hui Liu}

\authormark{Wang et al.}

\address[1]{\orgdiv{School of Computer Science and Engineering}, \orgname{Central South University},\orgaddress{\postcode{410075}, \state{Changsha}, \country{China}}}
\address[2]{\orgdiv{School of Computer Science and Technology}, \orgname{Nanjing Tech University}, \orgaddress{ \postcode{211816}, \state{Nanjing}, \country{China}}}
\address[3]{\orgdiv{School of Software}, \orgname{Xinjiang University}, \orgaddress{\postcode{830008}, \state{Urumqi}, \country{China}}}

\corresp[$\ast$]{Corresponding author. \href{email:leideng@csu.edu.cn}{leideng@csu.edu.cn}; \href{email:hliu@njtech.edu.cn}{hliu@njtech.edu.cn}}

\received{Date}{0}{Year}
\revised{Date}{0}{Year}
\accepted{Date}{0}{Year}

%\editor{Associate Editor: Name}

%\abstract{
%\textbf{Motivation:} .\\
%\textbf{Results:} .\\
%\textbf{Availability:} .\\
%\textbf{Contact:} \href{name@bio.com}{name@bio.com}\\
%\textbf{Supplementary information:} Supplementary data are available at \textit{Briefings in Bioinformatics}
%online.}

\abstract{\textbf{Motivation:} Drug combination therapy has become a increasingly promising method in the treatment of cancer. However, the number of possible drug combinations is so huge that it is hard to screen synergistic drug combinations through wet-lab experiments. Therefore, computational screening has become an important way to prioritize drug combinations. Graph neural network have recently shown remarkable performance in the prediction of compound-protein interactions, but it has not been applied to the screening of drug combinations. \\
\textbf{Results:} In this paper, we proposed a deep learning model based on graph neural networks and attention mechanism to identify drug combinations that can effectively inhibit the viability of specific cancer cells. The feature embeddings of drug molecule structure and gene expression profiles were taken as input to multi-layer feedforward neural network to identify the synergistic drug combinations. We compared DeepDDS with classical machine learning methods and other deep learning-based methods on benchmark data set, and the leave-one-out experimental results showed that DeepDDS achieved better performance than competitive methods. Also, on an independent test set released by well-known pharmaceutical enterprise AstraZeneca, DeepDDS was superior to competitive methods by more than 16\% predictive precision. Furthermore, we explored the interpretability of the graph attention network, and found the correlation matrix of atomic features revealed important chemical substructures of drugs. We believed that DeepDDS is an effective tool that prioritized synergistic drug combinations for further wet-lab experiment validation.\\
\textbf{Availability and implementation: } Source code and data are available at \href{https://github.com/Sinwang404/DeepDDs/tree/master}{https://github.com/Sinwang404/DeepDDS/tree/master} \\}
\keywords{Drug combination, attention mechanism, synergistic effect, graph neural network, deep learning, chemical structure}

% \boxedtext{
% \begin{itemize}
% \item Key boxed text here.
% \item Key boxed text here.
% \item Key boxed text here.
% \end{itemize}}

\maketitle

\section{Introduction}
Both traditional and modern medicine has taken advantage of the combined use of several active agents to treat diseases. Compared with single-drug therapy, the drug combinations often improve efficacy (\citealp{csermely2013structure}), reduce side effects (\citealp{zhao2013systems}) and overcome drug resistance (\citealp{hill2013genetic,verderosa11high}). Drug combinations are increasingly used to treat a variety of complex diseases, such as hypertension (\citealp{giles2014efficacy}), infectious diseases (\citealp{zheng2018drug}), and cancer (\citealp{kim2021anticancer}; \citealp{vitiello2021vulnerability}). For example, triple-negative breast cancer is a malignant tumor with strong invasiveness, high metastasis rate and poor prognosis. Lapatinib or Rapamycin alone has little therapeutic effect, but their combined treatment has been reported to significantly increase the apoptosis rate of triple-negative breast cancer cells (\citealp{liu2011combinatorial}). However, some drug combinations may cause antagonistic effect and even aggravate the disease (\citealp{azam2021trends}). Therefore, it is crucial to accurately discover synergistic drug combinations to specific diseases.

Traditional discovery of drug combinations is mainly based on clinical trials and limited to only a few number of drugs (\citealp{li2015large}), far from meeting the urgent need for anticancer drugs. Due to the great number of possible drug combinations, traditional method is cost-consuming and impractical. With the development of high-throughput drug screening technology, people can simultaneously carry out large-scale screening of drug combinations over hundreds of cancer cell lines (\citealp{hertzberg2000high}, \citealp{bajorath2002integration,macarron2011impact}). Torres \textit{et al.} utilized yeast to screen a large number of drug combinations and provided a method to identify preferential drug combinations for further testing in human cells (\citealp{torres2013high}). In despite of high degree of genomic correlation between the original tumor and the derived cancer cell line, \textit{in vitro} experiments of high-throughput drug screening still cannot accurately capture the  mode of action of drug molecules \textit{in vivo} (\citealp{ferreira2013importance}). Microcalorimetry screening (\citealp{kragh2021effective}) and genetically encoded fluorescent sensors  (\citealp{potekhina2021drug}) have been developed to screen effective antimicrobial combinations for \textit{in vivo} disease treatment. However,  these techniques require skilled operations and complicated experimental procedures.

In recent years, the datasets of single drug sensitivities to cancer cell lines increase greatly, such as Cancer Cell Line Encyclopedia (CCLE) (\citealp{barretina2012cancer}) and Genomics of Drug Sensitivity in Cancer (GDSC), which contains drug sensitivities to hundreds of human cancer cell lines, as well as gene expression profiles, mutants and copy number variants. Meanwhile, several large-scale data resource of drug combinations have been released. For example, O'Neil \textit{et al.} released a large-scale drug pair synergy study, which included more than 20,000 pairwise synergy scores between 38 unique drugs (\citealp{o2016unbiased}). The famous pharmaceutical company AstraZeneca (\citealp{menden2019community}) released their drug pair collaboration experiments, which includes 11,576 experiments of 910 drug combinations to 85 cancer cell lines with genome-related information. DrugCombDB (\citealp{liu2020drugcombdb}) has collected more than 6,000,000 quantitative drug dose responses, by which they calculated synergy scores to evaluate synergy or antagonism for each drug combination. In addition, quite a few data portal designed to collect drug combinations and relevant knowledge have been developed. The release of above-mentioned data resources motivated the development of computational screening of drug combinations. Many studies have been proposed to explore the vast space of drug combinations to identify synergistic efficacy. For example, classical machine learning methods, such as support vector machine (SVM) and random forest, successfully predicted the maximal antiallodynic effect of a new derivative of dihydrofuran-2-one (LPP1) used in combination with pregabalin (PGB) in the streptozocin-induced neuropathic pain model in mice (\citealp{salat2013application}, \citealp{qi2012random}).

Recently, the deep learning is increasingly applied to drug development and discovery. For example, DeepSynergy (\citealp{preuer2018deepsynergy}) combined the chemical information of drugs and genomic features of cancer cells to predict drug pairs with synergistic effects. TranSynergy (\citealp{liu2021transynergy}) is a mechanism-driven and self-attention boosted deep learning model that integrates information from gene-gene interaction networks, gene dependencies, and drug-target associations to predict synergistic drug combinations and deconvolute the cellular mechanisms. On the other hand, some studies applied SMILES to characterize chemical properties of drugs. For example, Gao et al. used the drug descriptors based on the SMILES to predict drug synergy. Liu et al. regarded the SMILES code as a string and directly input into a convolutional neural network (\citealp{liu2016drug}) to extract drug features for subsequent prediction task. More interesting, graph neural network is used to learn feature representation from drug chemical structure (\citealp{wu2018moleculenet,xiong2019pushing}).

In this paper, we propose a deep learning model, DeepDDS (Deep Learning for Drug-Drug Synergy prediction), to predict the synergistic effect of drug combinations. First, the drug chemical structure is represented by a graph in which the vertices are atoms and the edges are chemical bonds. Next, a graph convolutional network and attention mechanism is used to compute the drug embedding vectors. By integration of the genomic and pharmaceutical features, DeepDDS can capture important information from drug chemical structure and gene expression patterns to identify synergistic drug combinations to specific cancer cell lines. We compare DeepDDS to both classical machine learning methods (SVM, RF, GTB and XGBoost) and other latest deep learning (DTF, DeepSynergy and TranSynergy) on benchmark data set, DeepDDS significantly outperform other competitive methods. In particular, we conducted leave-one-out experiments to verify that DeepDDS achieved better performance when one drug (combination) or one tissue is not included in the training set. Also, on an independent test set released by well-known pharmaceutical enterprise AstraZeneca, DeepDDS was superior to competitive methods by more than 16\% predictive precision. We also explored the function of graph attention network in revealing important chemical substructures of drugs, and found the correlation matrix of atomic features showed clustering patterns among atom subgroups during the training process. Finally, we use the trained model to predict novel drug combination and find two previously reported synergistic drug combinations in the top 10 predicted results, MK2206 and AZD5363, MK2206 and AZD6244 to HCC1806 breast cancer cells. In summary, we believed that DeepDDS is an effective tool that prioritized synergistic drug combinations for further wet-lab experiment validation.

\section{Materials and methods}
\subsection{Data source}
The SMILES (Simplified Molecular Input Line Entry System) (\citealp{weininger1988smiles}) of drugs are obtained from DrugBank (\citep{wishart2018drugbank}), based on which the chemical structure of a drug can be converted to a graph using RDKit (\citep{landrum2006rdkit}).  In the molecular graph, the vertices are atoms and the edges are chemical bonds.

The gene expression data of cancer cell lines are obtained from Cancer Cell Line Encyclopedia (CCLE, ~\citealp{barretina2012cancer}), which is an independent project that makes effort to characterize genomes, mRNA expression, and anti-cancer drug dose responses across cancer cell lines. The expression data is normalized through TPM (Transcripts Per Million) based on the genome-wide read counts matrix.

To construct the benchmark set, we obtain the drug combination sensitivity data from a recently released large-scale oncology screening data set (\citealp{o2016unbiased}), where the viability of 39 cancer cells treated with thousands of drug combinations was evaluated by biochemical assay. The Loewe Additivity score (\citealp{loewe1953problem}), a quantitative metric that defines the synergistic or antagonistic effect of the drug combination, was calculated based on the 4 by 4 dose-response matrix using the Combenefit tool (\citealp{di2016combenefit}). Of note, multiple replicates of one drug combination were assayed in the original data, and thus the average score of the replicates was selected as the final synergistic score for each unique drug-pair-cell-line combination. According to the Loewe score, a combination with the score above zero is regarded as synergistic, and with the score below zero is antagonistic. Obviously, the drug combinations with higher synergistic scores are more attractive candidates for further clinical experiments. Since many additive combinations may exist (synergy scores are around 0 due to noise), we choose a stricter threshold to classify the combinations. Particularly, combinations with synergy score higher than 10 are labeled as positive (synergistic), and those with score less than 0 are labeled as negative (antagonistic). This yielded a balanced benchmark set that contains 12,415 unique drug pair-cell line combinations, covering 36 anticancer drugs and 31 human cancer cell lines.

\subsection{Pipeline of DeepDDS}
\begin{figure*}[htb]
	\centering
	\includegraphics[scale=0.5]{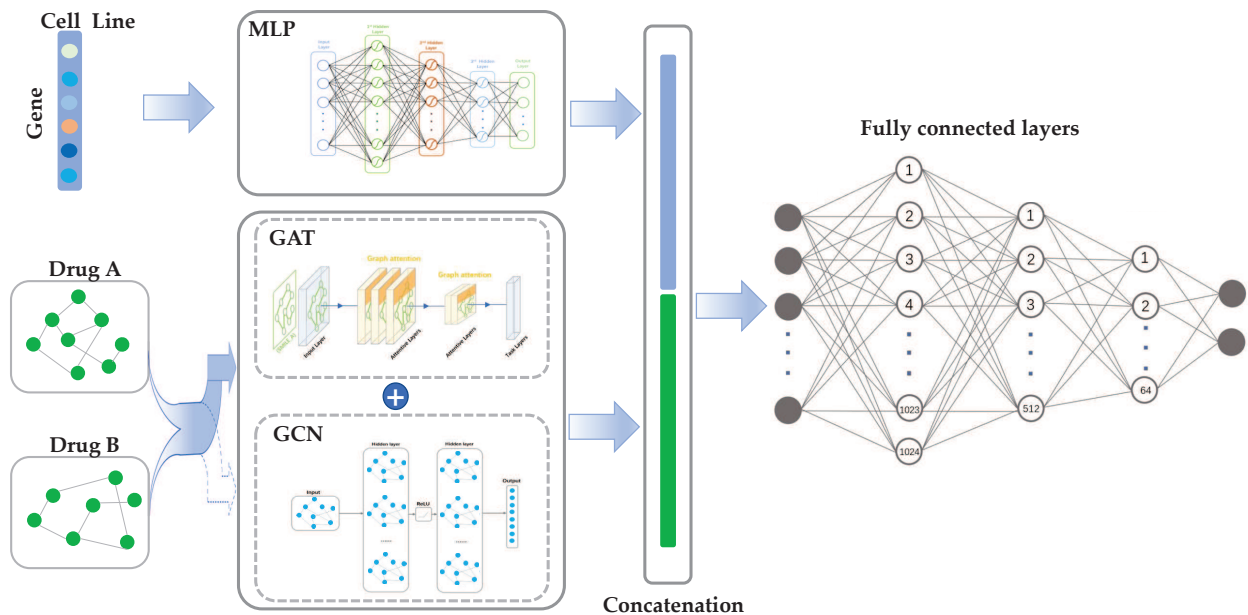}
	\caption{The pipeline of DeepDDS learning framework. The feature embedding of gene expression profiles of cancer cell line is obtained through Multi-Layer Perception (MLP), and the feature embedding of drug is obtained through GAT or GCN based on the drug molecular graph generated from drug SMILES. The embedding vectors of drug and cell line are subsequently concatenated to feed into a multi-layer fully connected network to predict the synergistic effect.}
	\label{DeepDDS}
\end{figure*}

Figure~\ref{DeepDDS} illustrates the end-to-end learning framework for the prediction of synergistic drug combinations. For each pairwise drug combination, the input layer firstly receives the molecular graphs of two drugs and gene expression profiles of one cancer cell line that was treated by these two drugs. We tested two type of Graph Neural Networks (GNN), graph attention network (GAT) and graph convolution network (GCN), to extract features of drugs. The genomic feature representation of cancer cells is encoded by a multi-layer perception (MLP). The embedding vectors are subsequently concatenated as the final feature representation of each drug-pair-cell-line combination, which is propagated through the fully-connected layers for the binary classification of drug combinations
(synergistic or antagonistic).

\subsection{Drug Representation based on GNN}
We use the open-source chemical informatics software RDKit (\citealp{landrum2006rdkit}) to convert the SMILES into molecular graphs, where the nodes are atoms and the edges are chemical bonds. Specifically, a graph for a given drug is defined as $G=(V,E)$, where $V$ is the set of $N$ nodes that represented by a $C$-dimensional vector, and $E$ is the set of edges represented as an adjacency matrix $A$. In a molecule graph, $x_{i}\in V$ is the $i$-th atom and $e_{ij}\in E$ is the chemical bond between the $i$-th and $j$-th atoms. The chemical molecular graph is non-Euclidean data and lacks of translation invariance, therefore, we applied graph neural network instead of traditional convolution network, to extract drug feature representations based on the graphs.

For each node in a graph, we use DeepChem (\citealp{ramsundar2019deep}) to compute a set of atomic attributes as its initial feature. Specifically, each node is represented as a binary vector including five pieces of information: the atom symbol, the number of adjacent atoms, the number of adjacent hydrogen, the implicit value of the atom, and whether the atom is in an aromatic structure. In GNN, the learning process of drug representation is actually the message passing between each node and its neighbor nodes. In this paper, we consider two types of GNN (graph convolution network and graph attention network) in our learning framework and evaluate their performance in the drug feature extraction.

\subsubsection{Graph Convolutional Network (GCN)}
The input of the multi-layer GCN is the node feature matrix $X \in \mathbb{R}^{N \times C} $ and the adjacency matrix $ A  \in \mathbb{R}^{N \times N} $ that represents the connection of nodes. According to Welling et al. (\citealp{kipf2016semi}), it can write dissemination rules in a standardized format to ensure stability.

The iteration process can be defined as below:
\begin{equation}
	\textit{H}^{(l+1)} = \sigma (\tilde{D}^{-\frac{1}{2}} \tilde{A} \tilde{D}^{-\frac{1}{2}} \textit{H}^{(l)} \textit{W}^{(l)})
	\label{eq:04}
	\vspace*{1pt}
\end{equation}
where $ \tilde{A} $ = $ \tilde{A} $ + $ I_{N} $($ I_{N} $ is the identity matrix) is the adjacency matrix of the undirected graph with added self-connections, $ \tilde{D}_{ii} = \sum_{i}\tilde{A}_{ii} $ ; $ \textit{H}^{(l+1)}  \in  \mathbb{R}^{N \times C} $ is the matrix of activation in the $l$th layer, $ \textit{H}^{(0)} = X $, $ \sigma $ is an activation function, and $ \textit{W} $ is a learnable parameter.

%approximate the layer-wise convolution operation, as illustrated in (\citealp{kipf2016semi}):
The output $ \textit{Z} \in \mathbb{R}^{N \times F} $ ($ F $ is the number of output features per node) can be defined as below:
\begin{equation}
	\textit{Z} = \tilde{D}^{-\frac{1}{2}} \tilde{A} \tilde{D}^{-\frac{1}{2}} \textit{X} \Theta
	\label{eq:03}
	\vspace*{1pt}
\end{equation}
where $ \Theta \in \mathbb{R}^{C \times F} $ ($ F $ is the number of filters or feature maps) is the matrix of filter parameters.

Our GCN-based model uses three consecutive GCN layers activated by ReLU function. The original GCN is a method for classify the node by semi-supervised learning, i.e., the outputs are the node-level feature vectors. To construct graph-level feature vectors, we use Sum, Average, and Max Pooling to aggregate the whole graph feature from learned node features and evaluate their performance. We find that the use of Max Pooling layer in GCN-based DeepDDS outperforms the others. Therefore, we add a global Max Pooling layer after the last GCN layer to extract the representation.

\subsubsection{Graph Attention Network (GAT)}
The graph attention network (GAT) proposes a multi-head attention-based architecture to learn higher-level features of nodes in a graph by applying a self-attention mechanism. Every attention head has its own parameters. The GAT architecture is built from the graphics attention layer. The output features for nodes were computed as

\begin{equation}
	h^{'}_{i} = ||_{m=1,...,M}(\alpha_{i,i}^{m} Wh_{i} + \sum_{j \in N(i)} \alpha_{i,j}^{m} Wh_{j})
	\label{eq:05}
	\vspace*{1pt}
\end{equation}

Where $ || $ concat the output results of multiple attention mechanisms, $ M $ is the number of attetion heads, and  $ W \in \mathbb{R}^{C^{\prime} \times C} $ is a weight matrix. The attention coefficient $ \alpha_{i,j} $, between each input node $ i $ and its first-order neighbor in the graph, is calculated as follows:
\begin{equation}
	\alpha_{i,j} = \dfrac{exp(elu(a^{T}[Wh_{i}||Wh_{j}]))}{\sum_{k \in N(i) } exp(elu(a^{T}[Wh_{i}||Wh_{k}]))}
	\label{eq:06}
	\vspace*{1pt}
\end{equation}
where $a^{T} \in \mathbb{R}^{C^{\prime}}$ is learnable weight vector, $ T $ is the corresponding transpose, and $ elu $ is a Non-linear activation function, when $ x $ is negative, $ y $ is equal to 0. Then, 'softmax' function is introduced to normalize all neighbor nodes $ j $ of $ i $ for easy calculation and comparison.

\subsection{Cell Line Feature Extraction based on MLP}
To alleviate the dimension imbalance between the feature vectors of drugs and cell lines, we selected the significant genes according to a LINCS project (\citealp{yang2012genomics}). The LINCS project provides a set of about 1000 carefully chosen genes, referred to as 'Landmark gene set', which can capture 80\% of the information based on the Connectivity Map (CMap) data (\citealp{cheng2016systematic}). The intersected genes between the CCLE gene expression profiles and the Landmark set was chosen for subsequent analysis. We used the gene annotation information in the Cancer Cell Line Encyclopedia (CCLE) (\citealp{barretina2012cancer}) and the GENCODE annotation database (\citealp{derrien2012gencode}) to remove the redundant data, as well as the transcripts of non-coding RNA. Finally, we select \textbf{954} genes from raw expression profiles as input to the model.

We adopt an MLP to extract the cell line features. The MLP includes two hidden layers, and the number of hidden units of each layer is selected via hyperparameter selection (See Hyperparameter setting for detail).

\subsection{Predicting the synergistic effect of drug combinations versus cell lines }
We formulated the prediction of synergistic drug combinations as an end-to-end binary classification model. Upon the embedding vectors of drugs through GAT or GCN, and the embedding vectors of cell lines through MLP, they are concatenated as the input of multiple fully-connected layers. We adopt the spindle-shaped structure for the fully connected layer. The probability of the synergistic effect (classification label) was computed by the softmax function that follows the output of the last hidden layer, as follows:
\begin{equation}
p_{t} = softmax \left ( W_{out}\cdot a^{l} +b_{out} \right )
\label{eq:07}
\vspace*{1pt}
\end{equation}
where $p_{t}$ is the probability of t, $W_{out}$ and $b_{out}$ are the weight matrix and bias vector, $ a^{l} $ are the embedding features learned by previous layers, as follows:
\begin{equation}
	a^{l} = \sigma (W^{l}a^{l-1} + b^{l})
	\label{eq:08}
	\vspace*{1pt}
\end{equation}
Where $l$ is the number of hidden layers, $W$ and $b$ are the matrices corresponding to all hidden layers and output layers, bias vector, $a^{0} =concat(R_{drug1},R_{drug2},R_{cellline})$ is the raw input vector.

Given a set of combinations with labels, we adopted the cross-entropy as the loss function to train the model, with the aim to minimize the loss during the training process, which is formulated as follows:
\begin{equation}
F=min\left ( -\sum_{i=1}^{N}log P_{t_{i}} +\frac{2}{\lambda }\left \| \Theta  \right \|\right )
\label{eq:09}
\vspace*{1pt}
\end{equation}
where $\Theta$ represents the set of all trainable weight and bias parameters involved in the model, $N$ is the total number of samples in the training dataset, $t_{i}$ is the $i$th label, and $\lambda$ is an L2 regularization hyper-parameter.

\section{Result}
\subsection{Hyperparameter setting}
The real architecture of DeepDDS is actually determined by hyperparameter setting. The hyperparameters cover the numbers of layers and units of each layer in MLP, GCN and GAN, as well as the activation function and learning rate. As exhaustive enumerations of the hyperparameters are computationally inhibitive, thereby we adopt grid-like search to tune the hyperparameters. As shown in Table~\ref{Tag_04}, we have tested different structural forms and values of these hyperparameters. We tuned the hyperparameters via five-fold cross validations on benchmark dataset. The selected values of these hyperparameters are displayed in boldface. The GCN yield to better performance in the drug feature extraction when its structure has three hidden layers and number of units are 1024, 512 and 156, respectively. We have also considered different number of hidden layers for GAN and MLP, and found they performed best with two hidden layers. For multi-head attention mechanism, multiple independent values are evaluated. For the activation function, the ELU and ReLU activation functions after the GAT layers at DeepDDS-GAT are used. For DeepDDS-GCN, it also has similar layer structure, but only ReLU is used as activation function.
\begin{table}[htb]
	\centering
	\caption{Hyperparameter settings of DeepDDS}
	\label{Tag_04}
	\setlength{\tabcolsep}{0.01mm}{
		\begin{tabular}{cc}
			\toprule
			\multicolumn{1}{c} {Hyperparameter}& \multicolumn{1}{c} {Values}\\ \hline
			GCN Hidden units & [1024,156];\textbf{[1024,512,156]};[512,256,156]\\
			GAN Hidden units & \textbf{[1024,512]};[512,128];[1024,156];[512,156]\\
			GAN attention Head & 4, 8, \textbf{10}, 12, 16 \\
			MLP Hidden units & [1024,512];\textbf{[2048,512]};[2048,1024];[4096,512]\\
			FC hidden units & [4096, 1024, 512]; [2048, 1024, 512]; \\
			&\textbf{[1024, 512, 128]}; [1024, 512, 64]\\
			Learning rate  & $10^{-2}$;$ \mathbf{10^{-3}}$; $10^{-4}$; $10^{-5}$ \\
			Dropout & No dropout,0.1; \textbf{0.2}; 0.3; 0.4; 0.5\\
			\botrule
	\end{tabular}}
\end{table}

\subsection{Performance comparison on cross-validation}
To evaluate the performance of DeepDDS, we compared DeepDDS with some current state-of-the-art methods, including both classical machine learning methods and deep learning-based methods. Six classical machine learning methods, including Random Forests (RF), Gradient Boosting Machines (GBM), Extreme Gradient Boosting (XGBoost), Adaboost, Multilayer Perceptron (MLP), Support Vector Machines (SVM), are considered in the performance comparison. Three deep learning-based methods are TranSynergy (\citealp{liu2021transynergy}), DeepSynergy (\citealp{preuer2018deepsynergy})  and Deep Tensor Factorization (\citealp{sun2020dtf}). To clarify the difference between DeepDDS and these deep learning-based methods, we summarize them as below:
\begin{itemize}	
	\item \textbf{TranSynergy.} TranSynergy includes three major components, input dimension reduction component, self-attention transformer component, and output fully connected component. It combines the network propagated drug target profile, gene dependency and gene expression to find novel genes associated with the synergistic drug combination from the learned biological relations.
	
	\item \textbf{DeepSynergy.} DeepSynergy uses molecular chemistry and cell line genomic information as input, and a cone layer in a neural network (DNN) to simulate drug synergy and finally predict the synergy score.
	
	\item \textbf{Deep Tensor Factorization (DTF).} DTF combine tensor-based framework and deep learning methods together to predict synergistic effect of drug pairs, which is comprised mainly by a tensor factorization method and a deep neural network.
\end{itemize}

First, we conducted five-fold cross validation to benchmark the predictive power of DeepDDS. The training samples (each sample is a drug-drug-cell line triplet) are randomly split into five subsets of roughly equal size, each subset is taken in turn as a test set and the remaining four subsets are used to train the model, whose prediction accuracy on the test set is then evaluated. The average prediction accuracy over the 5-folds is used as the final performance measure. For clarity, we provide typical performance measures widely used in  classification tasks, including area under the receiver operator characteristics curve (ROC AUC), area under the precision recall curve (PR AUC), accuracy (ACC), balanced accuracy (BACC), precision (PREC), sensitivity (TPR) and Cohen Kappa. Table \ref{Tag_01} shows these performance measures of DeepDDS and other methods. Clearly, DeepDDS-GAT achieved higher accuracy than all other methods, and its performance measures of ROC AUC, PR AUC, ACC, BACC, PREC, TPR, TNR and Kappa reach 0.93, 0.93, 0.85, 0.85, 0.85, 0.85, 0.85 and 0.71, respectively. In fact, both DeepDDS-GAT and DeepDDS-GCN outperform others in terms of all these performance measures. We note that the classifier XGBoost also achieved remarkable performance, nevertheless still inferior to DeepDDS. The three deep learning-based methods TranSynergy, DTF and DeepSynergy follow closely XGBoost, but outperform other methods. %Although DeepSynergy and DTF exhibited obtain relatively high ACC metrics, our model achieve outperforms other methods in other performance metrics.

\begin{table*}[htb]
	\centering
	\caption{Performance comparison of DeepDDS and competitive methods on 5-fold cross-validation}
	\label{Tag_01}
	\setlength{\tabcolsep}{2.5mm}{
		\begin{tabular}{llllllll}
			\toprule
			Performance Metric & ROC AUC & PR AUC & ACC & BACC & PREC & TPR & KAPPA \\
			\midrule
			DeepDDS-GAT & \textbf{0.93 $\pm$ 0.01}  & \textbf{0.93 $\pm$ 0.01} & \textbf{0.85 $\pm$ 0.07} & \textbf{0.85 $\pm$ 0.07} & \textbf{0.85 $\pm$ 0.07}  & \textbf{0.85 $\pm$ 0.07} & \textbf{0.71 $\pm$ 0.21} \\
			DeepDDS-GCN & \textbf{0.93 $\pm$ 0.01}  & \textbf{0.92 $\pm$ 0.01} & \textbf{0.85 $\pm$ 0.01} & \textbf{0.85 $\pm$ 0.01} & \textbf{0.85 $\pm$ 0.01}  & \textbf{0.84 $\pm$ 0.01} & \textbf{0.70 $\pm$ 0.22} \\
			XGBoost & 0.92 $\pm$ 0.01 & 0.92 $\pm$ 0.01 & 0.83 $\pm$ 0.01 & 0.83 $\pm$ 0.01 & 0.84 $\pm$ 0.01 & 0.84 $\pm$ 0.01 &  0.68 $\pm$ 0.01 \\
			Random Forest &  0.86 $\pm$ 0.02 &  0.85 $\pm$ 0.02 &  0.77 $\pm$ 0.01 &  0.77 $\pm$ 0.01 &  0.78 $\pm$ 0.02 & 0.74 $\pm$ 0.01 &  0.55 $\pm$ 0.04 \\
			GBM &  0.85 $\pm$ 0.02 &  0.85 $\pm$ 0.01 &  0.76 $\pm$ 0.02 &  0.76 $\pm$ 0.02 &  0.77 $\pm$ 0.01 & 0.74 $\pm$ 0.01 &  0.53 $\pm$ 0.04 \\
			Adaboost &  0.83 $\pm$ 0.01 &  0.83 $\pm$ 0.03 &  0.74 $\pm$ 0.01 &  0.74 $\pm$ 0.02 &  0.74 $\pm$ 0.02 & 0.72 $\pm$ 0.01 &  0.48 $\pm$ 0.03 \\
			MLP &  0.65 $\pm$ 0.02 &  0.63 $\pm$ 0.05 &  0.56 $\pm$ 0.06 &  0.56 $\pm$ 0.05 &  0.54 $\pm$ 0.04 & 0.53 $\pm$ 0.22 &  0.12 $\pm$ 0.04 \\
			SVM &  0.58 $\pm$ 0.01 &  0.56 $\pm$ 0.02 &  0.54 $\pm$ 0.01 &  0.54 $\pm$ 0.01 &  0.54 $\pm$ 0.01 & 0.51 $\pm$ 0.12 &  0.08 $\pm$ 0.04 \\
			TranSynergy & 0.90 $\pm$ 0.01 & 0.89 $\pm$ 0.01 & 0.83 $\pm$ 0.01 & 0.83 $\pm$ 0.01 & 0.84 $\pm$ 0.01 & 0.80 $\pm$ 0.01 &  0.64 $\pm$ 0.01 \\
			DTF & 0.89 $\pm$ 0.01 & 0.88 $\pm$ 0.01 & 0.81 $\pm$ 0.01 & 0.81 $\pm$ 0.01 & 0.82 $\pm$ 0.01 & 0.77 $\pm$ 0.03 &  0.63 $\pm$ 0.04 \\
			DeepSynergy & 0.88 $\pm$ 0.01 & 0.87 $\pm$ 0.01 & 0.80 $\pm$ 0.01 & 0.80 $\pm$ 0.01 & 0.81 $\pm$ 0.01 & 0.75 $\pm$ 0.01 &  0.59 $\pm$ 0.05 \\
			\botrule
	\end{tabular}}
\end{table*}

We further checked the top 100 drug pairs with highest predicted synergy scores by DeepDDS-GAT (For detail see Supplementary Table S1), and found that 98 drug pairs have been experimentally validated to be synergistic combinations over different cancer cell lines.

\subsection{Performance evaluation by input permutation}
\begin{figure}[htbp]  %figure 3
	\captionsetup{labelformat=simple, position=top}
	\centering
	\subfloat[]
	{
		\begin{minipage}[b]{.5\columnwidth}
			\centering
			\includegraphics[width=4.5cm]{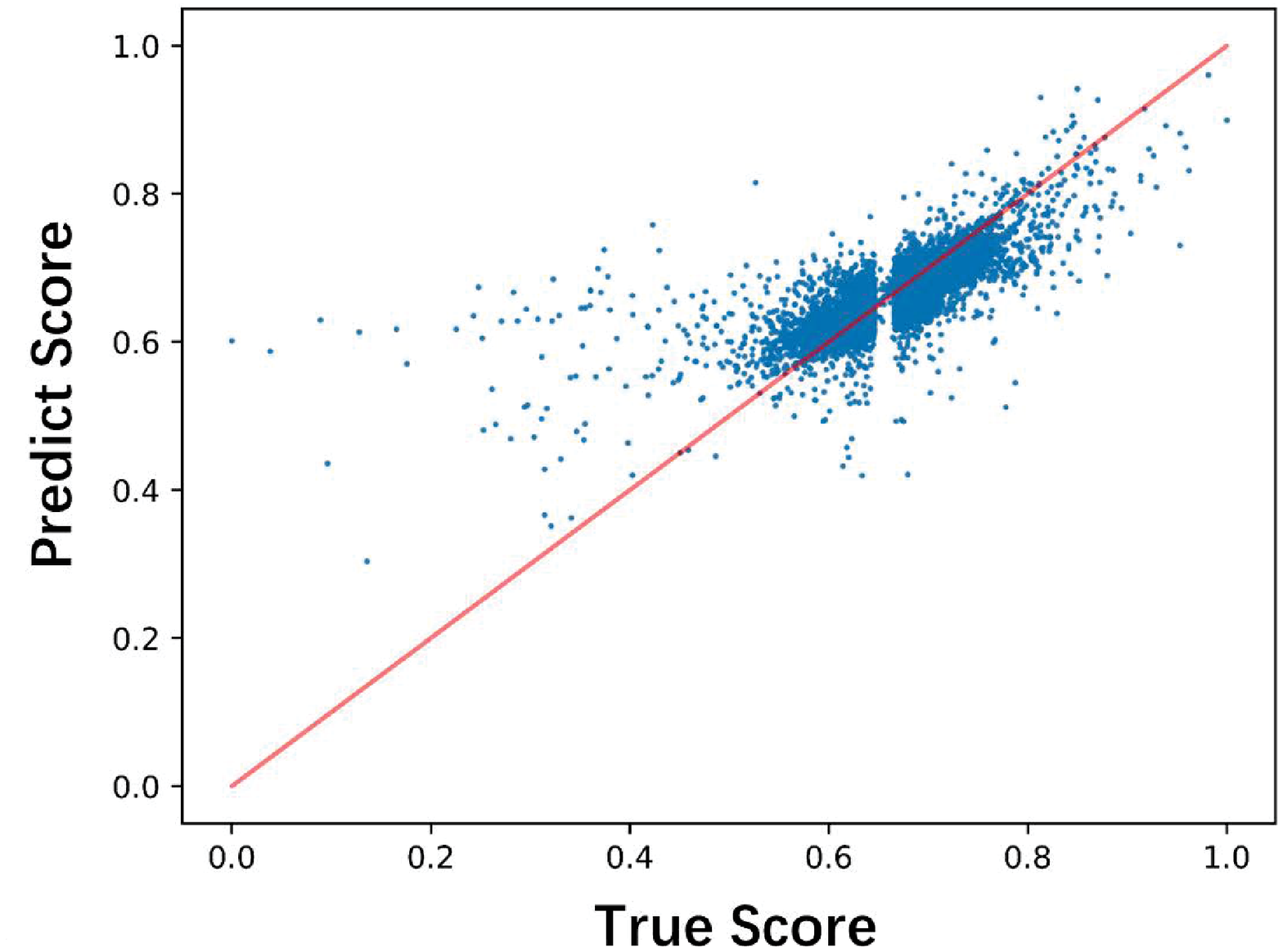}
	\end{minipage}}
	\subfloat[]
	{
		\begin{minipage}[b]{.5\columnwidth}
			\centering
			\includegraphics[width=4.5cm]{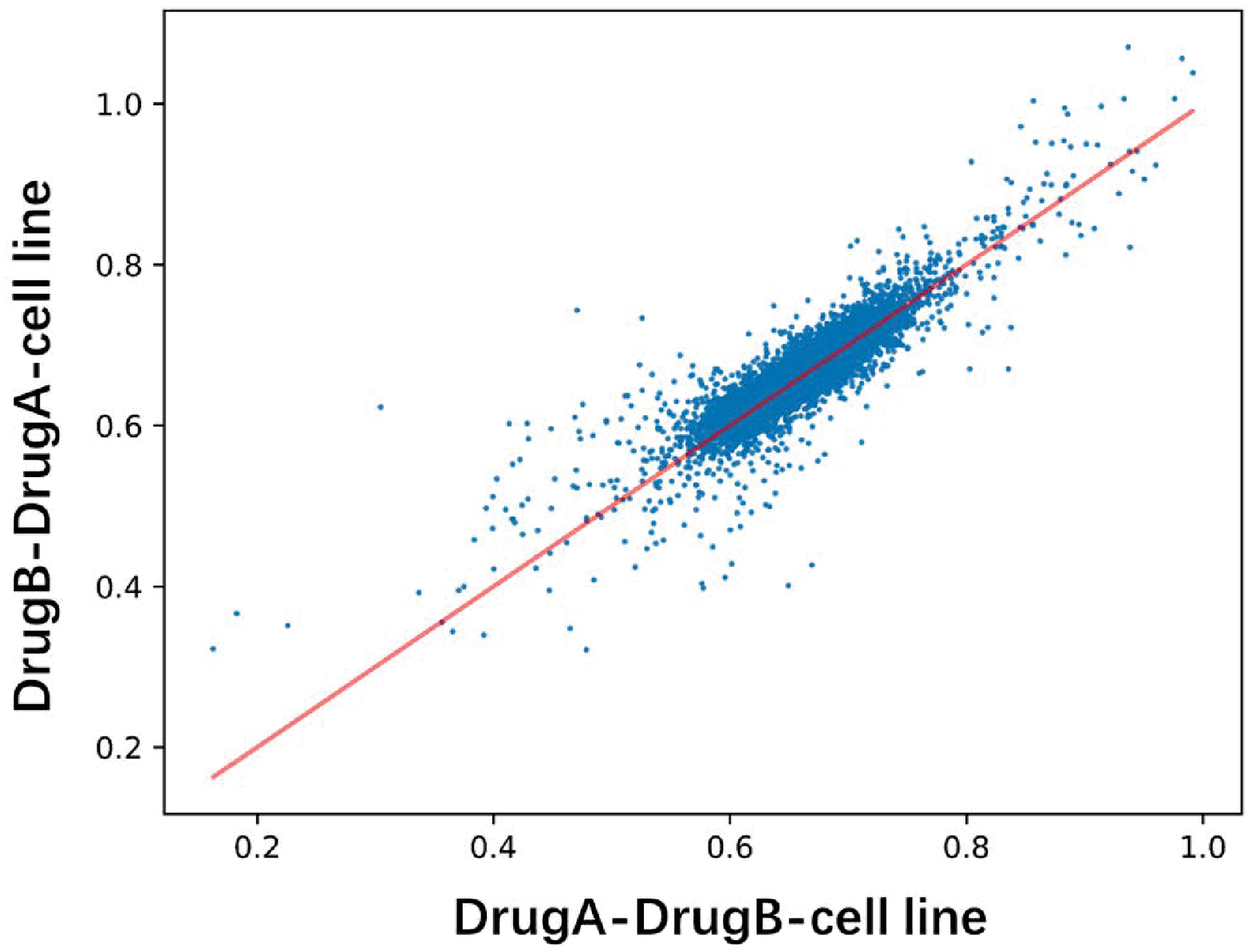}
	\end{minipage}} \\
	\caption{Scatter plots of synergy scores. (a) The scatter plot with respect to the real synergy scores and predicted synergy score. (b) The scatter plot of synergy score obtained from different input order of two drugs.}
	\label{fig:3}
\end{figure}

We found that the higher the real synergy score, the higher the predictive score. After normalization of real synergy scores to [0, 1] region, we draw a scatter plot of the drug combinations with respect to the predicted and real synergy scores. As shown in Figure~\ref{fig:3} (a), most points locate closely to the identity line. The Pearson correlation between the predicted synergy scores and real synergy scores reach 0.801. The results indicate our method achieve superior predictive accuracy.

We go further to verify the predictive performance of DeepDDS upon different input order of two drugs. For drug A and drug B, we permutate the input features so that drug A-drug B and drug B-drug A are regarded as two different samples to train the model. Figure \ref{fig:3} (b) shows the predicted synergy score upon different sequence of input features by DeepDDS-GAT. It can be found that most values locate closely to the identity line and the Pearson correlation coefficient reach 0.9. It can prove that our model is insensitive to the sequence of the input features of drug combinations. In addition, we found that the ROC AUC and PR AUC obtained by drug A-drug B and drug B-drug A both reach or be close to 0.93.
\subsection{Performance evaluation by leave-one-out cross validation}
We went further to verify the performance of the DeepDDS model using leave-one-out cross validation.
First, we conducted the leave-one drug combination-out experiment. More precisely, we iteratively exclude each drug combination from the training set, use the remaining data to train the DeepDDS model that is in turn used to predict the sensitivity of the excluded drug combination to cancer cell lines. The result of the leave-one drug combination-out experiment is shown in Table~\ref{Tag_03}, DeepDDS-GAT achieve notably performance by AUC value 0.89, followed by DeepDDS-GCN. It can be also found that DeepDDS significantly outperform all other  methods.

As the leave-one drug combination-out experiment did not excluded single drug from the training set, we next leave one drug out to prevent the information of certain drug being seen by the model. The leave-one drug-out experiment check the potential to learn the important features of unseen drug from the chemical structures of those seen drugs. As shown in Table~\ref{Tag_03}, DeepDDS still achieve better performance than other competitive methods.

\begin{figure}[htb]
	\centering
	\includegraphics[scale=0.25]{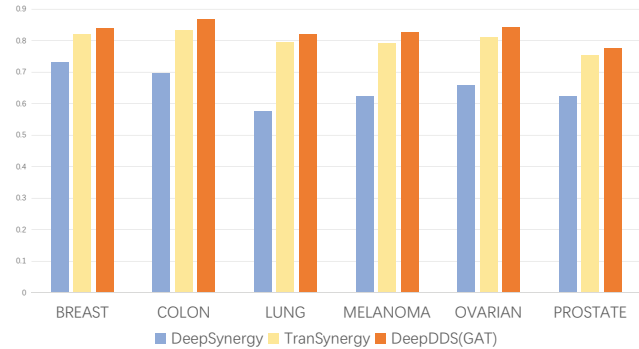}
	\caption{The ROC AUC values of DeepDDS-GAT, DeepSynergy and TranSynergy upon leave-tissue-out evaluations on six different tissues, including breast, colon, lung, melanoma, ovarian and prostate.}
	\label{leave_tissue_out}
\end{figure}

As previous studies (\citealp{liu2021transynergy}), we also carried out leave-one cell line-out experiment to verify the performance of DeepDDS. Take the cell line T47D as an example, the drug combination between BEZ-235 and MK-8669, Dasatinib, Lapatinib, Geldanamycin, PD325901, Erlotinib, MK-4541, Temozolomide, Vinorelbine, ABT-888, all have a high experimental synergy scores (Loewe$>$100). Expectedly, the prediction scores of these drug combinations have prior rankings among all candidate drug pairs (See Supplementary Table S2-S3 for detail).  In addition to the leave-one cell line-out evaluation , \citealp{preuer2018deepsynergy}), we adopt more rigorous strategy to evaluate our method. We exclude all the cancer cell lines belong to specific tissue from the training set, so that the model can not see any gene expression information of a certain type of tissue. We iteratively use the excluded cancer cell lines as the validation set and the remaining samples as the training set to train the model. Table~\ref{Tag_03} illustrated that DeepDDS-GAT achieve the best performance on leave-one tissue-out evaluation. Also, DeepDDS performs better than all classical machine learning methods and deep learning-based methods. Moreover, Figure~\ref{leave_tissue_out} show the ROC AUC values of DeepDDS-GAT, DeepSynergy and TranSynergy on six different tissues, including breast, colon, lung, melanoma, ovarian and prostate. It can be found that DeepDDS-GAT is better than other two deep learning-based methods with ROC AUC 0.84, 0.867, 0.821, 0.828, 0.843 and 0.775 by leave-one tissue-out cross validation, respectively.

\begin{table*}[htb]
	\centering
	\caption{Performance on DeepDDS and competitive methods on leave-drug combination-out, leave-drug-out and leave-tissue-out experiments}
	\label{Tag_03}
	\setlength{\tabcolsep}{1mm}{
		\begin{tabular}{cccccccccc}
			\toprule
			& \multicolumn{3}{c} {Leave-drug combination-out}  & \multicolumn{3}{c} {Leave-drug-out} & \multicolumn{3}{c} {Leave-tissue-out}\\
			\midrule
			& ROC AUC   & PR AUC  & ACC & ROC AUC   & PR AUC  & ACC & ROC AUC   & PR AUC  & ACC
			\\
			DeepDDS-GAT & \textbf{0.89 $\pm$ 0.02}  & \textbf{0.88 $\pm$ 0.06} & \textbf{0.81 $\pm$ 0.03} & \textbf{0.73 $\pm$ 0.01} & \textbf{0.72 $\pm$ 0.05}  & \textbf{0.66 $\pm$ 0.02} & \textbf{0.83 $\pm$ 0.04} & \textbf{0.82 $\pm$ 0.4} &  \textbf{0.74 $\pm$ 0.03} \\
			XGBoost & 0.84 $\pm$ 0.02 & 0.83 $\pm$ 0.04 & 0.75 $\pm$ 0.02 & 0.66 $\pm$ 0.09 & 0.65 $\pm$ 0.06 & 0.61 $\pm$ 0.06 &  0.82 $\pm$ 0.01 & 0.81 $\pm$ 0.01 &  0.73 $\pm$ 0.01\\
			TranSynergy &  &  &  &  &  &  &  0.81 $\pm$ 0.01 & 0.79 $\pm$ 0.02 &  0.73 $\pm$ 0.03 \\
			DeepSynergy & 0.83 $\pm$ 0.03 & 0.81 $\pm$ 0.05 & 0.77 $\pm$ 0.03 & 0.71 $\pm$ 0.07 & 0.64 $\pm$ 0.06 & 0.61 $\pm$ 0.07 &  0.80 $\pm$ 0.01 & 0.79 $\pm$ 0.04 &  0.71 $\pm$ 0.05\\
			Random Forest &  0.82 $\pm$ 0.02 &  0.81 $\pm$ 0.03 &  0.74 $\pm$ 0.02 &  0.67 $\pm$ 0.08 &  0.66 $\pm$ 0.05 & 0.62 $\pm$ 0.06 &  0.80 $\pm$ 0.08 & 0.80 $\pm$ 0.05 &  0.71 $\pm$ 0.05 \\
			MLP &  0.82 $\pm$ 0.03 &  0.81 $\pm$ 0.05 &  0.74 $\pm$ 0.02 &  0.69 $\pm$ 0.05 &  0.68 $\pm$ 0.04 & 0.62$\pm$ 0.06 &  0.77 $\pm$ 0.07 & 0.76 $\pm$ 0.05 &  0.70 $\pm$ 0.06\\
			GBM &  0.81 $\pm$ 0.03 &  0.81 $\pm$ 0.04 &  0.74 $\pm$ 0.02 &  0.64 $\pm$ 0.09 &  0.63 $\pm$ 0.09 & 0.60 $\pm$ 0.06 &  0.81 $\pm$ 0.08 & 0.81 $\pm$ 0.05 &  0.72 $\pm$ 0.06\\
			Adaboost &  0.77 $\pm$ 0.02 &  0.78 $\pm$ 0.02 &  0.69 $\pm$ 0.03 &  0.62 $\pm$ 0.11 &  0.61 $\pm$ 0.06 & 0.58 $\pm$ 0.11 &  0.77 $\pm$ 0.12 & 0.78 $\pm$ 0.11 &  0.70 $\pm$ 0.11\\
			SVM &  0.66 $\pm$ 0.01 &  0.65 $\pm$ 0.05 &  0.58 $\pm$ 0.01 &  0.60 $\pm$ 0.02 &  0.59 $\pm$ 0.05 & 0.55 $\pm$ 0.03 &  0.66 $\pm$ 0.04 & 0.66 $\pm$ 0.07 &  0.59 $\pm$ 0.05\\
			\botrule
	\end{tabular}}
\end{table*}

\subsection{Evaluation on independent test set}
To verify the generalization ability of our method, we use the benchmark dataset  (\citealp{o2016unbiased}) to train our model, and then employ an independent test set released by AstraZeneca (\citealp{menden2019community}) to evaluate the performance of DeepDDS and other competitive methods. The independent test set contains 668 unique drug-pair-cell-line combinations, covering 57 drugs (Supplementary Table S4) and 24 cell lines (Supplementary Table S5).

\begin{figure}[htb]
	\centering
	\includegraphics[scale=0.8]{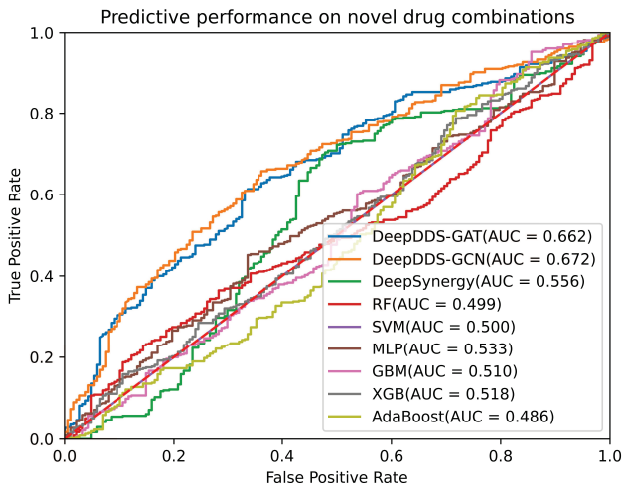}
	\caption{ROC curves and AUC values of DeepDDS and competitive methods on independent test dataset released by AstraZeneca.}
	\label{roc_indep}
\end{figure}

Table \ref{Tag_02} shows the performance achieved by DeepDDS and competitive methods on the independent test set. It can be seen that the performance of DeepDDS is better than all competitive methods in terms of every performance measure. For clarity, we draw the ROC curves of DeepDDS and other methods, as shown in Figure~\ref{roc_indep}. DeepDDS-GAT and DeepDDS-GCN account for top 2, followed by DeepSynergy. Meanwhile, it can be found that most machine learning-based methods perform just as random guess. This result indicate classical machine learning methods run into overfitting, while deep learning-based method acquire better generalization ability. In particular, DeepDDS-GAT and DeepDDS-GCN correctly predicted 421 (421/668=0.63) and 402 (402/668=0.6) drug pairs included in the independent test set, which outperform DeepSynergy correct prediction 317 (317/668=0.47) by 16\% and 13\%, respectively. The confusion matrices in Figure S3 show detailed numbers of correctly and falsedly predicted samples by the three methods.

\begin{table*}[htb]
	\centering
	\caption{Performance metrics for the classification task in independent test set}
	\label{Tag_02}
	\setlength{\tabcolsep}{2.5mm}{
		\begin{tabular}{llllllll}
			\toprule
			Performance Metric & ROC AUC & PR AUC & ACC & BACC & PREC & TPR & KAPPA \\
			\midrule
			DeepDDS-GAT & 0.66 $\pm$ 0.12  & 0.82 $\pm$ 0.15 & \textbf{0.64 $\pm$ 0.15} & 0.62 $\pm$ 0.13 & 0.80 $\pm$ 0.11  & \textbf{0.67 $\pm$ 0.12} & \textbf{0.21 $\pm$ 0.29} \\
			DeepDDS-GCN & \textbf{0.67 $\pm$ 0.12}  & \textbf{0.83 $\pm$ 0.13} & 0.60 $\pm$ 0.11 & \textbf{0.63 $\pm$ 0.13} & \textbf{0.83 $\pm$ 0.10}  & 0.56 $\pm$ 0.20 & 0.21 $\pm$ 0.23 \\
			%	TranSynergy & 0.59 $\pm$ 0.13 & 0.76 $\pm$ 0.14 & 0.53 $\pm$ 0.12 & 0.57 $\pm$ 0.13 & 0.79 $\pm$ 0.11 & 0.47 $\pm$ 0.19 &  0.13 $\pm$ 0.17 \\
			DeepSynergy & 0.55 $\pm$ 0.15 & 0.71 $\pm$ 0.13 & 0.47 $\pm$ 0.14 & 0.53 $\pm$ 0.13 & 0.75 $\pm$ 0.14 & 0.39 $\pm$ 0.17 &  0.04 $\pm$ 0.15 \\
			Random Forest &  0.53 $\pm$ 0.14 &  0.76 $\pm$ 0.16 &  0.50 $\pm$ 0.14 &  0.54 $\pm$ 0.13 &  0.75 $\pm$ 0.14 & 0.49 $\pm$ 0.14 &  0.06 $\pm$ 0.11 \\
			MLP & 0.53 $\pm$ 0.13 & 0.74 $\pm$ 0.12 & 0.53 $\pm$ 0.15 & 0.53 $\pm$ 0.15 & 0.74 $\pm$ 0.13 & 0.53 $\pm$ 0.13 &  0.05 $\pm$ 0.11 \\
			GBM &  0.51 $\pm$ 0.10 &  0.71 $\pm$ 0.09 &  0.45 $\pm$ 0.12 &  0.47 $\pm$ 0.08 &  0.69 $\pm$ 0.14 & 0.43 $\pm$ 0.12 &  -0.03 $\pm$ 0.14 \\
			XGBoost &  0.52 $\pm$ 0.11 &  0.73 $\pm$ 0.12 &  0.45 $\pm$ 0.15 &  0.49 $\pm$ 0.11 &  0.71 $\pm$ 0.09 & 0.38 $\pm$ 0.17 &  -0.01 $\pm$ 0.14 \\
			Adaboost &  0.49 $\pm$ 0.09 &  0.69 $\pm$ 0.14 &  0.46 $\pm$ 0.17 &  0.47 $\pm$ 0.12 &  0.69 $\pm$ 0.14 & 0.46 $\pm$ 0.15 &  -0.05 $\pm$ 0.17 \\
			SVM &  0.47 $\pm$ 0.11 &  0.71 $\pm$ 0.13 &  0.54 $\pm$ 0.13 &  0.47 $\pm$ 0.15 &  0.70 $\pm$ 0.13 & 0.63 $\pm$ 0.11 &  -0.04 $\pm$ 0.15 \\
			\botrule
	\end{tabular}}
\end{table*}

\subsection{Graph attention network reveals important chemical substructure}
DeepDDS-GAT model iteratively passes messages between nodes so that each node can capture the information of its neighboring nodes. Meanwhile, each neuron is connected to the neighborhood upper layer through a set of learnable weights in the GAT network. As a result, the feature representation actually encodes the information of the chemical substructure around the atom, including formal charge, water solubility and other physicochemical properties. This motivate us to explore the implications of the attention mechanism in revealing the important chemical substructures.

\begin{figure*}[htb]
	\centering
	\includegraphics[scale=0.4]{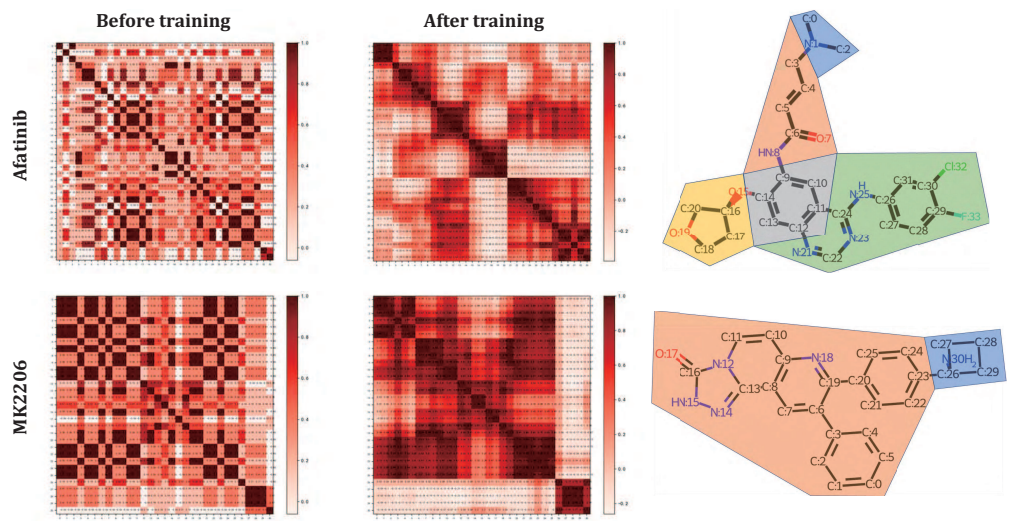}
	\caption{Heat map of the atomic feature similarity matrices of Afatinib and MK2206 before and after training. The heat maps show clear clustering patterns during the learning process. The diagrams of chemical structures of Afatinib and MK2206 display the five and two subgroups according to their clusters of heat maps. }
	\label{atomSM}
\end{figure*}

For example, previous study has showed that EGFR inhibitor Afatinib and AKT inhibitor MK2206 play synergistic effect in the treatment of lung cancer, head and neck squamous cell carcinoma (HNSCC)   (\citealp{modjtahedi2014comprehensive},\citealp{silva2017akt},\citealp{hung2016epidermal}).  We investigate how the atomic feature vectors evolved during the learning process, by measuring the Pearson correlation coefficient between atom pairs based on the feature vectors. The heat maps of the atom correlation matrix is plotted to observe the change of feature patterns. The similarity scores are displayed in the cells and indicated by the color scheme. It can be seen that before training the visual patterns in the heat maps of two drugs shows some degree of chaos. After training, however, the heat map of both drugs show obvious atomic clusters in a specific order. In particular, the atom of drug Afatinib is clustered into five subgroups, while MK2206 clustered into two atom subgroups (one big and one small block), as shown in figure~\ref{atomSM}. Without loss of generalization, we randomly select a few other drug combinations to check whether their feature vectors undergo similar pattern changes during the training process. These drug combinations include AZD2014 and AZD6244, AZD8931 and AZD5363, GDC0941 and AZD6244,GDC0941 and MK2206. As expected, the atomic feature vectors of the involved drugs gradually cluster into several subgroups (See Supplementary Figure S2-S6 for more detail).

\begin{figure}[htbp]  %figure 5
	\captionsetup{labelformat=simple, position=top}
	\centering
	\subfloat[]
	{
		\begin{minipage}[b]{\linewidth}
			\centering
			\includegraphics[width=8cm]{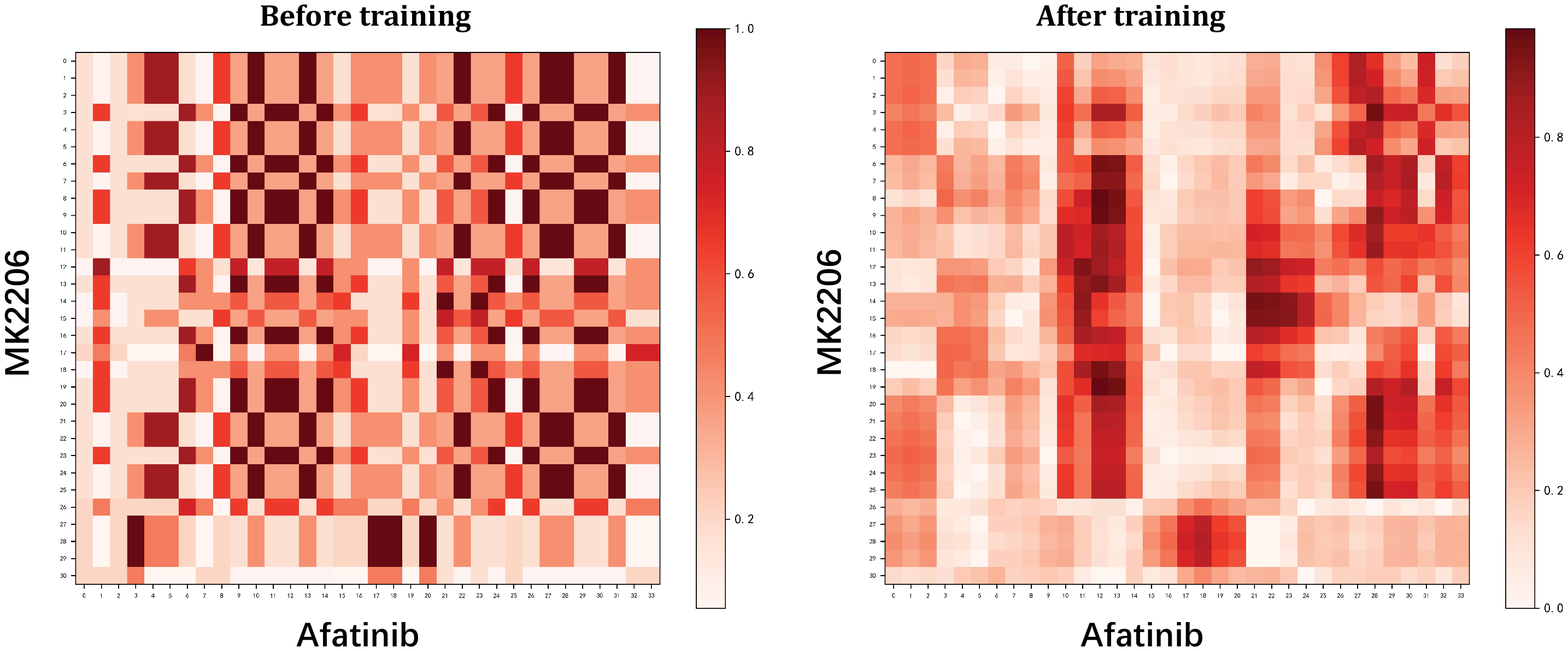}
	\end{minipage}} \\
	\subfloat[]
	{
		\begin{minipage}[b]{\linewidth}
			\centering
			\includegraphics[width=8cm]{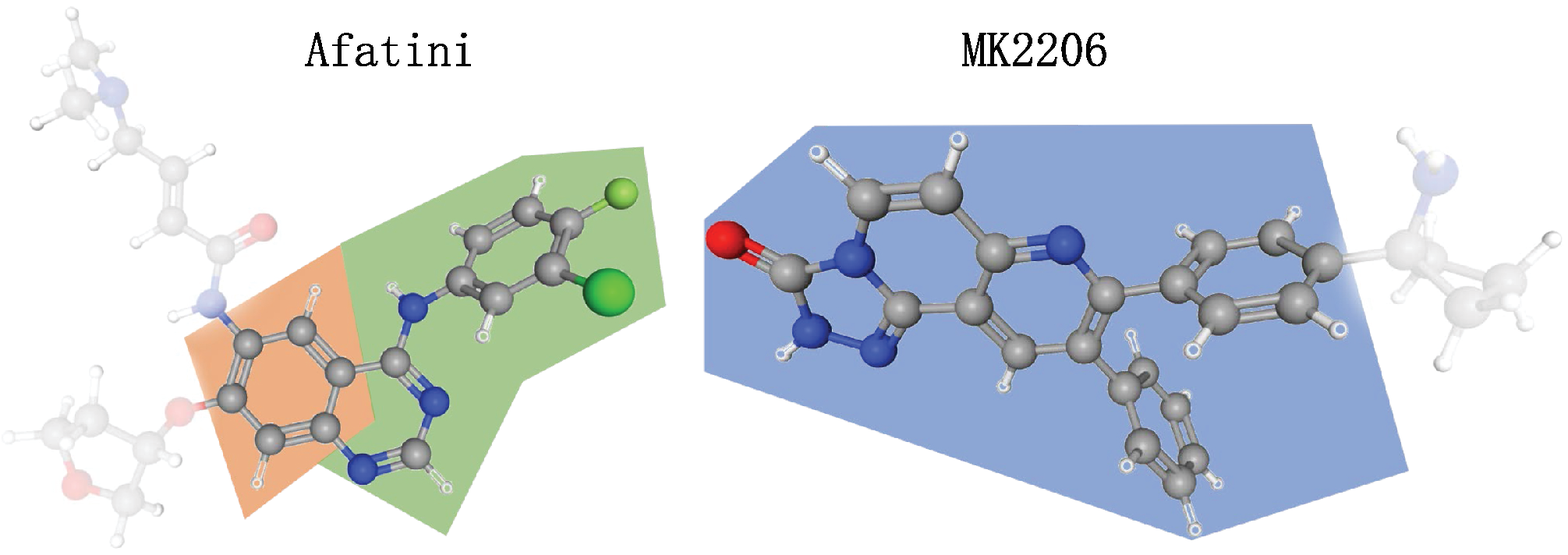}
	\end{minipage}} \\
	\caption{The heat maps of Pearson correlation coefficients between atom pairs across Afatinib and MK2206. Pearson correlation coefficients are computed using the feature vector before and after training.  (a) The heat map shows no clear visual pattern before training, but after training shows two clear linking blocks. (b) The bigger atom subgroup (No.1-25 atoms) of MK2206 associates remarkably to the 3th and 5th atom subgroups (No.9-14 atoms and No.21-33 atoms) of Afatinib.}
	\label{comb_drug}
\end{figure}

We go a step further to explore the interpretability of the graph attention network in revealing the chemical substructures that are potential components exerting synergistic effect of the drug combinations. We compute the Pearson correlation coefficients between atom pairs across two drugs, so that significant association between chemical subgroups of different drugs can be uncovered. Take the drug combinations Afatinib and MK2206 as example again, we find that the heat map of the atom correlation matrix have no clear clustering pattern before training, while it show two notably linking blocks after training, as shown in Figure~\ref{comb_drug} (a). More interesting, these two linking blocks exactly indicate that the bigger atom subgroup (No.1-25 atoms) of MK2206 associates to the 3th and 5th atom subgroups (No.9-14 atoms and No.21-33 atoms) of Afatinib. From the 3D structures of the two drugs, they are just the main functional groups of Afatinib and MK2206, respectively. For other examples mentioned above, we found that their inter-drug atom correlation matrices also display clustering patterns, as shown in Figure S3-S6.

As a result, the atom embedding vectors display clear feature patterns during the training process, namely, the atom correlation matrices clearly cluster into several atom subgroups, and the degree of association between atom subgroups of different drugs transfer from chaos to order. We adventure to speculate that the atom subgroups included in these two drugs play key role in their synergistic function, although the pharmacological mechanism in vivo remains unclear to date.

\subsection{Predicting novel synergistic combinations}
%The performance evaluation experiments above have shown that our DeepDDS model achieve superior performance, so that we apply DeepDDS to predict novel synergistic combination.  We use the O'Neil drug combination dataset to train the model, and selected 76,608 candidate drug pairs generated from AstraZeneca dataset (\citealp{menden2019community}) for discovery exploration.

The performance evaluation experiments above have shown that our DeepDDS model achieve superior performance, thereby we apply DeepDDS to predict novel synergistic combination. We use the O'Neil drug combination dataset to train the DeepDDS model. To generate candidate drug combinations, we selected 42 small molecule targeted drugs approved by the FDA (\citealp{bedard2020small}) and then generated 855 candidate drug pairs (see Supplementary Table S6). We listed the top 10 predicted drug combinations in Table~\ref{novel_drugs}. To verify the reliability of the predicted results, we conducted an non-exhaustive literature search and found there are at least six predicted drug combinations are consistent with the observations in previous studies or under clinical trials. We presented the PMIDs or DOI identifiers of these related publications in Table~\ref{novel_drugs}.

\begin{table}[htb]
	\centering
	\caption{Top 10 predicted novel synergistic combinations on A375 cancer cell line}
	\label{novel_drugs}
	\setlength{\tabcolsep}{0.1mm}{
		\begin{tabular}{p{0.23\columnwidth}<{\raggedright}p{0.22\columnwidth}<{\raggedright}p{0.1\columnwidth}<{\raggedright}p{0.13\columnwidth}<{\raggedright}p{0.38\columnwidth}<{\raggedright}}
			\toprule
			Drug A & Drug B & Cell line & Predict Score &  Publications\\
			\midrule
			\textbf{Abemaciclib} & \textbf{Lapatinib}  & A375 & 0.9977 & 26977878, 33389550, 26977873		\\
			Binimetinib & Sorafenib  & A375 & 0.9974 & NA \\
			Copanlisib& Regorafenib  & A375 & 0.9973 & NA \\
			\textbf{Copanlisib}& \textbf{Sorafenib}  & A375 & 0.9973 &30962952, 27259258, doi:10.5282/edoc.24304 \\
			Binimetinib& Regorafenib  & A375 & 0.9971 & NA \\
			\textbf{Erlotinib}& \textbf{Regorafenib}  & A375 & 0.997
			&	25907508 \\
			\textbf{Vemurafenib}& \textbf{Sorafenib} & A375 & 0.9969 & 33119140, 30076136, 30844744, 29605720, doi:10.21037/tcr.2020.01.62 \\
			Vemurafenib&Regorafenib  & A375 & 0.9967 & NA \\
			\textbf{Lapatinib}& \textbf{Regorafenib}  & A375 & 0.9967 & 27864115, 24911215 \\
			Pazopanib& Sorafenib  & A375 & 0.9965 & NA \\

			\botrule
	\end{tabular}}
\end{table}

For example, the CDK4/6 inhibitor \textbf{abemaciclib} and  EGFR inhibitor \textbf{lapatinib}  significantly enhanced growth-inhibitory for HER2-positive breast cancer (\citealp{goel2016overcoming}). Ye \textit{et al.} found that \textbf{Copanlisib} reduced \textbf{Sorafenib}-induced phosphorylation of p-AKT and enhanced synergistically of antineoplastic effect in vitro (\citealp{ye2019pi3k}). Also, the combination of \textbf{Erlotinib} and \textbf{Regorafenib} in the treatment of hepatocellular carcinoma successfully overcome the interference of epidermal growth factors (\citealp{d2015modulation}). Addition of \textbf{Sorafenib} to \textbf{Vemurafenib} increased ROS production through ferroptosis, thus increasing the sensitivity of melanoma cells to vemurafenib (\citealp{tang2020sorafenib}).
Zhang et al. reproted that the \textbf{Regorafenib} combined with the \textbf{Lapatinib} could improve anti-tumor efficacy in human colorectal cancer (\citealp{zhang2017synergistic}). We believe that other predicted drug pairs are also promising combinations await for further validation.

\section{Discussion and Conclusion}
In this paper, we have proposed a novel method to predict synergistic drug combination to specific cancer cells.  Overall, our method performs significantly better than other competitive methods on the five-fold cross validation experiments. However, we noticed that the predictive accuracy of our method are still limited on the independent test set, although the performance of our method is greatly superior than all competitive methods. We think the limited performance is mainly attributed to the small number of training samples. In fact, the benchmark dataset actually includes only 38 unique drugs and 39 cancer cell lines, while the space for possible drug combinations is much larger when novel drugs are included

Two different graph neural network, GAT and GCN, are used to learn drug embedding vectors in our method. We extensively compared their performance to each other, as well as to quite a few competitive methods. Overall, GAT performs slightly better than GCN, and thus we further explored the interpretability of the GAT model. However, we have realized that the physicochemical properties of the molecular graph and attention weights between the atoms have not been fully understood. In the future, we are interested in studying the connections between atoms to incorporate more information resources into the DeepDDS model to improve the model interpretability and predictability.

In conclusion, we have proposed a novel method DeepDDS to predict the synergy of drug combinations for cancer cell lines with high accuracy. Our performance comparison experiments showed that DeepDDS performs better than other competitive methods. We have demonstrate that DeepDDS achieve state-of-the-art performance in a cross-validation setting with an independent test set. We believe that with the increasing size of the data set available, DeepDDS can be further improved and applied to other fields where drug combinations play an essential role, such as antiviral (\citealp{akhtar2020covid19}), antifungal (\citealp{pereira2021vitro}) and multi-drug synergy prediction(\citealp{ontong2021synergistic}). Overall, we believe that our method will yield some inspiring insights into the discovery of synergistic drug combinations.

\section{Competing interests}
There is NO Competing Interest.

\section{Author contributions statement}
J.W. and H.L. conceived the main idea and the framework of the manuscript. J.W. drafted the manuscript. J.W. and X. L. collected the data and performed the experiments. L.D. and H.L. helped to improve the idea and the manuscript.	S.S. reviewed drafts of the paper. L.D. and H.L. supervised the study and provided funding. All authors read and commented on the manuscript.

\section{Acknowledgments}
This work was supported by the National Natural Science Foundation of China under grants No.~61972422 and No.~62072058.

%USE THE BELOW OPTIONS IN CASE YOU NEED AUTHOR YEAR FORMAT.
\bibliographystyle{abbrvnat}
\bibliography{reference}

\begin{thebibliography}{52}
\providecommand{\natexlab}[1]{#1}
\providecommand{\url}[1]{\texttt{#1}}
\expandafter\ifx\csname urlstyle\endcsname\relax
  \providecommand{\doi}[1]{doi: #1}\else
  \providecommand{\doi}{doi: \begingroup \urlstyle{rm}\Url}\fi

\bibitem[Akhtar(2020)]{akhtar2020covid19}
M.~J. Akhtar.
\newblock Covid19 inhibitors: a prospective therapeutics.
\newblock \emph{Bioorg. Chem}, 101:\penalty0 104027, 2020.

\bibitem[Azam and Vazquez(2021)]{azam2021trends}
F.~Azam and A.~Vazquez.
\newblock Trends in phase ii trials for cancer therapies.
\newblock \emph{Cancers}, 13\penalty0 (2):\penalty0 178, 2021.

\bibitem[Bajorath(2002)]{bajorath2002integration}
J.~Bajorath.
\newblock Integration of virtual and high-throughput screening.
\newblock \emph{Nature Reviews Drug Discovery}, 1\penalty0 (11):\penalty0
  882--894, 2002.

\bibitem[Barretina et~al.(2012)Barretina, Caponigro, Stransky, Venkatesan,
  Margolin, Kim, Wilson, Leh{\'a}r, Kryukov, Sonkin,
  et~al.]{barretina2012cancer}
J.~Barretina, G.~Caponigro, N.~Stransky, K.~Venkatesan, A.~A. Margolin, S.~Kim,
  C.~J. Wilson, J.~Leh{\'a}r, G.~V. Kryukov, D.~Sonkin, et~al.
\newblock The cancer cell line encyclopedia enables predictive modelling of
  anticancer drug sensitivity.
\newblock \emph{Nature}, 483\penalty0 (7391):\penalty0 603--607, 2012.

\bibitem[Bedard et~al.(2020)Bedard, Hyman, Davids, and Siu]{bedard2020small}
P.~L. Bedard, D.~M. Hyman, M.~S. Davids, and L.~L. Siu.
\newblock Small molecules, big impact: 20 years of targeted therapy in
  oncology.
\newblock \emph{The Lancet}, 395\penalty0 (10229):\penalty0 1078--1088, 2020.

\bibitem[Cheng and Li(2016)]{cheng2016systematic}
L.~Cheng and L.~Li.
\newblock Systematic quality control analysis of lincs data.
\newblock \emph{CPT: pharmacometrics \& systems pharmacology}, 5\penalty0
  (11):\penalty0 588--598, 2016.

\bibitem[Csermely et~al.(2013)Csermely, Korcsm{\'a}ros, Kiss, London, and
  Nussinov]{csermely2013structure}
P.~Csermely, T.~Korcsm{\'a}ros, H.~J. Kiss, G.~London, and R.~Nussinov.
\newblock Structure and dynamics of molecular networks: a novel paradigm of
  drug discovery: a comprehensive review.
\newblock \emph{Pharmacology \& therapeutics}, 138\penalty0 (3):\penalty0
  333--408, 2013.

\bibitem[D’Alessandro et~al.(2015)D’Alessandro, Refolo, Lippolis, Carella,
  Messa, Cavallini, and Carr]{d2015modulation}
R.~D’Alessandro, M.~G. Refolo, C.~Lippolis, N.~Carella, C.~Messa,
  A.~Cavallini, and B.~I. Carr.
\newblock Modulation of regorafenib effects on hcc cell lines by epidermal
  growth factor.
\newblock \emph{Cancer chemotherapy and pharmacology}, 75\penalty0
  (6):\penalty0 1237--1245, 2015.

\bibitem[Derrien et~al.(2012)Derrien, Johnson, Bussotti, Tanzer, Djebali,
  Tilgner, Guernec, Martin, Merkel, Knowles, et~al.]{derrien2012gencode}
T.~Derrien, R.~Johnson, G.~Bussotti, A.~Tanzer, S.~Djebali, H.~Tilgner,
  G.~Guernec, D.~Martin, A.~Merkel, D.~G. Knowles, et~al.
\newblock The gencode v7 catalog of human long noncoding rnas: analysis of
  their gene structure, evolution, and expression.
\newblock \emph{Genome research}, 22\penalty0 (9):\penalty0 1775--1789, 2012.

\bibitem[Di~Veroli et~al.(2016)Di~Veroli, Fornari, Wang, Mollard, Bramhall,
  Richards, and Jodrell]{di2016combenefit}
G.~Y. Di~Veroli, C.~Fornari, D.~Wang, S.~Mollard, J.~L. Bramhall, F.~M.
  Richards, and D.~I. Jodrell.
\newblock Combenefit: an interactive platform for the analysis and
  visualization of drug combinations.
\newblock \emph{Bioinformatics}, 32\penalty0 (18):\penalty0 2866--2868, 2016.

\bibitem[Ferreira et~al.(2013)Ferreira, Adega, and
  Chaves]{ferreira2013importance}
D.~Ferreira, F.~Adega, and R.~Chaves.
\newblock The importance of cancer cell lines as in vitro models in cancer
  methylome analysis and anticancer drugs testing.
\newblock \emph{Oncogenomics and cancer proteomics-novel approaches in
  biomarkers discovery and therapeutic targets in cancer}, pages 139--166,
  2013.

\bibitem[Giles et~al.(2014)Giles, Weber, Basile, Gradman, Bharucha, Chen,
  Pattathil, Investigators, et~al.]{giles2014efficacy}
T.~D. Giles, M.~A. Weber, J.~Basile, A.~H. Gradman, D.~B. Bharucha, W.~Chen,
  M.~Pattathil, N.-M.-.~S. Investigators, et~al.
\newblock Efficacy and safety of nebivolol and valsartan as fixed-dose
  combination in hypertension: a randomised, multicentre study.
\newblock \emph{The Lancet}, 383\penalty0 (9932):\penalty0 1889--1898, 2014.

\bibitem[Goel et~al.(2016)Goel, Wang, Watt, Tolaney, Dillon, Li, Ramm, Palmer,
  Yuzugullu, Varadan, et~al.]{goel2016overcoming}
S.~Goel, Q.~Wang, A.~C. Watt, S.~M. Tolaney, D.~A. Dillon, W.~Li, S.~Ramm,
  A.~C. Palmer, H.~Yuzugullu, V.~Varadan, et~al.
\newblock Overcoming therapeutic resistance in her2-positive breast cancers
  with cdk4/6 inhibitors.
\newblock \emph{Cancer cell}, 29\penalty0 (3):\penalty0 255--269, 2016.

\bibitem[Hertzberg and Pope(2000)]{hertzberg2000high}
R.~P. Hertzberg and A.~J. Pope.
\newblock High-throughput screening: new technology for the 21st century.
\newblock \emph{Current opinion in chemical biology}, 4\penalty0 (4):\penalty0
  445--451, 2000.

\bibitem[Hill et~al.(2013)Hill, Ammar, Torti, Nislow, and
  Cowen]{hill2013genetic}
J.~A. Hill, R.~Ammar, D.~Torti, C.~Nislow, and L.~E. Cowen.
\newblock Genetic and genomic architecture of the evolution of resistance to
  antifungal drug combinations.
\newblock \emph{PLoS Genet}, 9\penalty0 (4):\penalty0 e1003390, 2013.

\bibitem[Hung et~al.(2016)Hung, Chen, Lung, Lin, Li, and
  Tsai]{hung2016epidermal}
M.-S. Hung, I.-C. Chen, J.-H. Lung, P.-Y. Lin, Y.-C. Li, and Y.-H. Tsai.
\newblock Epidermal growth factor receptor mutation enhances expression of
  cadherin-5 in lung cancer cells.
\newblock \emph{PLoS One}, 11\penalty0 (6):\penalty0 e0158395, 2016.

\bibitem[Kim et~al.(2021)Kim, Zheng, Tang, Jim~Zheng, Li, and
  Jiang]{kim2021anticancer}
Y.~Kim, S.~Zheng, J.~Tang, W.~Jim~Zheng, Z.~Li, and X.~Jiang.
\newblock Anticancer drug synergy prediction in understudied tissues using
  transfer learning.
\newblock \emph{Journal of the American Medical Informatics Association},
  28\penalty0 (1):\penalty0 42--51, 2021.

\bibitem[Kipf and Welling(2016)]{kipf2016semi}
T.~N. Kipf and M.~Welling.
\newblock Semi-supervised classification with graph convolutional networks.
\newblock \emph{arXiv preprint arXiv:1609.02907}, 2016.

\bibitem[Kragh et~al.(2021)Kragh, Gij{\'o}n, Maruri, Antonelli, Coppi, Kolpen,
  Crone, Tellapragada, Hasan, Radmer, et~al.]{kragh2021effective}
K.~N. Kragh, D.~Gij{\'o}n, A.~Maruri, A.~Antonelli, M.~Coppi, M.~Kolpen,
  S.~Crone, C.~Tellapragada, B.~Hasan, S.~Radmer, et~al.
\newblock Effective antimicrobial combination in vivo treatment predicted with
  microcalorimetry screening.
\newblock \emph{Journal of Antimicrobial Chemotherapy}, 2021.

\bibitem[Landrum et~al.(2006)]{landrum2006rdkit}
G.~Landrum et~al.
\newblock Rdkit: Open-source cheminformatics.
\newblock 2006.

\bibitem[Li et~al.(2015)Li, Huang, Fu, Wang, Wu, Ru, Zheng, Guo, Chen, Zhou,
  et~al.]{li2015large}
P.~Li, C.~Huang, Y.~Fu, J.~Wang, Z.~Wu, J.~Ru, C.~Zheng, Z.~Guo, X.~Chen,
  W.~Zhou, et~al.
\newblock Large-scale exploration and analysis of drug combinations.
\newblock \emph{Bioinformatics}, 31\penalty0 (12):\penalty0 2007--2016, 2015.

\bibitem[Liu et~al.(2020)Liu, Zhang, Zou, Wang, Deng, and
  Deng]{liu2020drugcombdb}
H.~Liu, W.~Zhang, B.~Zou, J.~Wang, Y.~Deng, and L.~Deng.
\newblock Drugcombdb: a comprehensive database of drug combinations toward the
  discovery of combinatorial therapy.
\newblock \emph{Nucleic acids research}, 48\penalty0 (D1):\penalty0 D871--D881,
  2020.

\bibitem[Liu and Xie(2021)]{liu2021transynergy}
Q.~Liu and L.~Xie.
\newblock Transynergy: Mechanism-driven interpretable deep neural network for
  the synergistic prediction and pathway deconvolution of drug combinations.
\newblock \emph{PLoS computational biology}, 17\penalty0 (2):\penalty0
  e1008653, 2021.

\bibitem[Liu et~al.(2016)Liu, Tang, Chen, and Wang]{liu2016drug}
S.~Liu, B.~Tang, Q.~Chen, and X.~Wang.
\newblock Drug-drug interaction extraction via convolutional neural networks.
\newblock \emph{Computational and mathematical methods in medicine}, 2016,
  2016.

\bibitem[Liu et~al.(2011)Liu, Yacoub, Taliaferro-Smith, Sun, Graham, Dolan,
  Lobo, Tighiouart, Yang, Adams, et~al.]{liu2011combinatorial}
T.~Liu, R.~Yacoub, L.~D. Taliaferro-Smith, S.-Y. Sun, T.~R. Graham, R.~Dolan,
  C.~Lobo, M.~Tighiouart, L.~Yang, A.~Adams, et~al.
\newblock Combinatorial effects of lapatinib and rapamycin in triple-negative
  breast cancer cells.
\newblock \emph{Molecular cancer therapeutics}, 10\penalty0 (8):\penalty0
  1460--1469, 2011.

\bibitem[Loewe(1953)]{loewe1953problem}
S.~Loewe.
\newblock The problem of synergism and antagonism of combined drugs.
\newblock \emph{Arzneimittelforschung}, 3:\penalty0 285--290, 1953.

\bibitem[Macarron et~al.(2011)Macarron, Banks, Bojanic, Burns, Cirovic,
  Garyantes, Green, Hertzberg, Janzen, Paslay, et~al.]{macarron2011impact}
R.~Macarron, M.~N. Banks, D.~Bojanic, D.~J. Burns, D.~A. Cirovic, T.~Garyantes,
  D.~V. Green, R.~P. Hertzberg, W.~P. Janzen, J.~W. Paslay, et~al.
\newblock Impact of high-throughput screening in biomedical research.
\newblock \emph{Nature reviews Drug discovery}, 10\penalty0 (3):\penalty0
  188--195, 2011.

\bibitem[Menden et~al.(2019)Menden, Wang, Mason, Szalai, Bulusu, Guan, Yu,
  Kang, Jeon, Wolfinger, et~al.]{menden2019community}
M.~P. Menden, D.~Wang, M.~J. Mason, B.~Szalai, K.~C. Bulusu, Y.~Guan, T.~Yu,
  J.~Kang, M.~Jeon, R.~Wolfinger, et~al.
\newblock Community assessment to advance computational prediction of cancer
  drug combinations in a pharmacogenomic screen.
\newblock \emph{Nature communications}, 10\penalty0 (1):\penalty0 1--17, 2019.

\bibitem[Modjtahedi et~al.(2014)Modjtahedi, Cho, Michel, and
  Solca]{modjtahedi2014comprehensive}
H.~Modjtahedi, B.~C. Cho, M.~C. Michel, and F.~Solca.
\newblock A comprehensive review of the preclinical efficacy profile of the
  erbb family blocker afatinib in cancer.
\newblock \emph{Naunyn-Schmiedeberg's archives of pharmacology}, 387\penalty0
  (6):\penalty0 505--521, 2014.

\bibitem[O'Neil et~al.(2016)O'Neil, Benita, Feldman, Chenard, Roberts, Liu, Li,
  Kral, Lejnine, Loboda, et~al.]{o2016unbiased}
J.~O'Neil, Y.~Benita, I.~Feldman, M.~Chenard, B.~Roberts, Y.~Liu, J.~Li,
  A.~Kral, S.~Lejnine, A.~Loboda, et~al.
\newblock An unbiased oncology compound screen to identify novel combination
  strategies.
\newblock \emph{Molecular cancer therapeutics}, 15\penalty0 (6):\penalty0
  1155--1162, 2016.

\bibitem[Ontong et~al.(2021)Ontong, Ozioma, Voravuthikunchai, and
  Chusri]{ontong2021synergistic}
J.~C. Ontong, N.~F. Ozioma, S.~P. Voravuthikunchai, and S.~Chusri.
\newblock Synergistic antibacterial effects of colistin in combination with
  aminoglycoside, carbapenems, cephalosporins, fluoroquinolones, tetracyclines,
  fosfomycin, and piperacillin on multidrug resistant klebsiella pneumoniae
  isolates.
\newblock \emph{Plos one}, 16\penalty0 (1):\penalty0 e0244673, 2021.

\bibitem[Pereira et~al.(2021)Pereira, de~Menezes, de~Oliveira, de~Oliveira, and
  Scorzoni]{pereira2021vitro}
T.~C. Pereira, R.~T. de~Menezes, H.~C. de~Oliveira, L.~D. de~Oliveira, and
  L.~Scorzoni.
\newblock In vitro synergistic effects of fluoxetine and paroxetine in
  combination with amphotericin b against cryptococcus neoformans.
\newblock \emph{Pathogens and Disease}, 2021.

\bibitem[Potekhina et~al.(2021)Potekhina, Bass, Kelmanson, Fetisova, Ivanenko,
  Belousov, and Bilan]{potekhina2021drug}
E.~S. Potekhina, D.~Y. Bass, I.~V. Kelmanson, E.~S. Fetisova, A.~V. Ivanenko,
  V.~V. Belousov, and D.~S. Bilan.
\newblock Drug screening with genetically encoded fluorescent sensors: Today
  and tomorrow.
\newblock \emph{International Journal of Molecular Sciences}, 22\penalty0
  (1):\penalty0 148, 2021.

\bibitem[Preuer et~al.(2018)Preuer, Lewis, Hochreiter, Bender, Bulusu, and
  Klambauer]{preuer2018deepsynergy}
K.~Preuer, R.~P. Lewis, S.~Hochreiter, A.~Bender, K.~C. Bulusu, and
  G.~Klambauer.
\newblock Deepsynergy: predicting anti-cancer drug synergy with deep learning.
\newblock \emph{Bioinformatics}, 34\penalty0 (9):\penalty0 1538--1546, 2018.

\bibitem[Qi(2012)]{qi2012random}
Y.~Qi.
\newblock Random forest for bioinformatics.
\newblock In \emph{Ensemble machine learning}, pages 307--323. Springer, 2012.

\bibitem[Ramsundar et~al.(2019)Ramsundar, Eastman, Walters, and
  Pande]{ramsundar2019deep}
B.~Ramsundar, P.~Eastman, P.~Walters, and V.~Pande.
\newblock \emph{Deep learning for the life sciences: applying deep learning to
  genomics, microscopy, drug discovery, and more}.
\newblock " O'Reilly Media, Inc.", 2019.

\bibitem[Sa{\l}at and Sa{\l}at(2013)]{salat2013application}
R.~Sa{\l}at and K.~Sa{\l}at.
\newblock The application of support vector regression for prediction of the
  antiallodynic effect of drug combinations in the mouse model of
  streptozocin-induced diabetic neuropathy.
\newblock \emph{Computer methods and programs in biomedicine}, 111\penalty0
  (2):\penalty0 330--337, 2013.

\bibitem[Silva-Oliveira et~al.(2017)Silva-Oliveira, Melendez, Martinho, Zanon,
  de~Souza~Viana, Carvalho, and Reis]{silva2017akt}
R.~J. Silva-Oliveira, M.~Melendez, O.~Martinho, M.~F. Zanon, L.~de~Souza~Viana,
  A.~L. Carvalho, and R.~M. Reis.
\newblock Akt can modulate the in vitro response of hnscc cells to irreversible
  egfr inhibitors.
\newblock \emph{Oncotarget}, 8\penalty0 (32):\penalty0 53288, 2017.

\bibitem[Sun et~al.(2020)Sun, Huang, Jiang, and Hu]{sun2020dtf}
Z.~Sun, S.~Huang, P.~Jiang, and P.~Hu.
\newblock Dtf: Deep tensor factorization for predicting anticancer drug
  synergy.
\newblock \emph{Bioinformatics}, 36\penalty0 (16):\penalty0 4483--4489, 2020.

\bibitem[Tang et~al.(2020)Tang, Li, Liu, Chen, and Han]{tang2020sorafenib}
F.~Tang, S.~Li, D.~Liu, J.~Chen, and C.~Han.
\newblock Sorafenib sensitizes melanoma cells to vemurafenib through
  ferroptosis.
\newblock \emph{Translational Cancer Research}, 9\penalty0 (3):\penalty0
  1584--+, 2020.

\bibitem[Torres et~al.(2013)Torres, Lee, Giaever, Nislow, and
  Brown]{torres2013high}
N.~P. Torres, A.~Y. Lee, G.~Giaever, C.~Nislow, and G.~W. Brown.
\newblock A high-throughput yeast assay identifies synergistic drug
  combinations.
\newblock \emph{Assay and drug development technologies}, 11\penalty0
  (5):\penalty0 299--307, 2013.

\bibitem[Verderosa et~al.()Verderosa, Dhouib, Hong, Anderson, Heras, and
  Totsika]{verderosa11high}
A.~D. Verderosa, R.~Dhouib, Y.~Hong, T.~K. Anderson, B.~Heras, and M.~Totsika.
\newblock A high-throughput cell-based assay pipeline for the preclinical
  development of bacterial dsba inhibitors as antivirulence therapeutics.
\newblock \emph{Scientific Reports}, 11\penalty0 (1):\penalty0 1--13.

\bibitem[Vitiello et~al.(2021)Vitiello, Martini, Mele, Giunta, De~Falco,
  Ciardiello, Belli, Cardone, Matrone, Poliero,
  et~al.]{vitiello2021vulnerability}
P.~P. Vitiello, G.~Martini, L.~Mele, E.~F. Giunta, V.~De~Falco, D.~Ciardiello,
  V.~Belli, C.~Cardone, N.~Matrone, L.~Poliero, et~al.
\newblock Vulnerability to low-dose combination of irinotecan and niraparib in
  atm-mutated colorectal cancer.
\newblock \emph{Journal of Experimental \& Clinical Cancer Research},
  40\penalty0 (1):\penalty0 1--15, 2021.

\bibitem[Weininger(1988)]{weininger1988smiles}
D.~Weininger.
\newblock Smiles, a chemical language and information system. 1. introduction
  to methodology and encoding rules.
\newblock \emph{Journal of chemical information and computer sciences},
  28\penalty0 (1):\penalty0 31--36, 1988.

\bibitem[Wishart et~al.(2018)Wishart, Feunang, Guo, Lo, Marcu, Grant, Sajed,
  Johnson, Li, Sayeeda, et~al.]{wishart2018drugbank}
D.~S. Wishart, Y.~D. Feunang, A.~C. Guo, E.~J. Lo, A.~Marcu, J.~R. Grant,
  T.~Sajed, D.~Johnson, C.~Li, Z.~Sayeeda, et~al.
\newblock Drugbank 5.0: a major update to the drugbank database for 2018.
\newblock \emph{Nucleic acids research}, 46\penalty0 (D1):\penalty0
  D1074--D1082, 2018.

\bibitem[Wu et~al.(2018)Wu, Ramsundar, Feinberg, Gomes, Geniesse, Pappu,
  Leswing, and Pande]{wu2018moleculenet}
Z.~Wu, B.~Ramsundar, E.~N. Feinberg, J.~Gomes, C.~Geniesse, A.~S. Pappu,
  K.~Leswing, and V.~Pande.
\newblock Moleculenet: a benchmark for molecular machine learning.
\newblock \emph{Chemical science}, 9\penalty0 (2):\penalty0 513--530, 2018.

\bibitem[Xiong et~al.(2019)Xiong, Wang, Liu, Zhong, Wan, Li, Li, Luo, Chen,
  Jiang, et~al.]{xiong2019pushing}
Z.~Xiong, D.~Wang, X.~Liu, F.~Zhong, X.~Wan, X.~Li, Z.~Li, X.~Luo, K.~Chen,
  H.~Jiang, et~al.
\newblock Pushing the boundaries of molecular representation for drug discovery
  with the graph attention mechanism.
\newblock \emph{Journal of medicinal chemistry}, 63\penalty0 (16):\penalty0
  8749--8760, 2019.

\bibitem[Yang et~al.(2012)Yang, Soares, Greninger, Edelman, Lightfoot, Forbes,
  Bindal, Beare, Smith, Thompson, et~al.]{yang2012genomics}
W.~Yang, J.~Soares, P.~Greninger, E.~J. Edelman, H.~Lightfoot, S.~Forbes,
  N.~Bindal, D.~Beare, J.~A. Smith, I.~R. Thompson, et~al.
\newblock Genomics of drug sensitivity in cancer (gdsc): a resource for
  therapeutic biomarker discovery in cancer cells.
\newblock \emph{Nucleic acids research}, 41\penalty0 (D1):\penalty0 D955--D961,
  2012.

\bibitem[Ye et~al.(2019)Ye, Mayerle, Ziesch, Reiter, Gerbes, and
  De~Toni]{ye2019pi3k}
L.~Ye, J.~Mayerle, A.~Ziesch, F.~P. Reiter, A.~L. Gerbes, and E.~N. De~Toni.
\newblock The pi3k inhibitor copanlisib synergizes with sorafenib to induce
  cell death in hepatocellular carcinoma.
\newblock \emph{Cell death discovery}, 5\penalty0 (1):\penalty0 1--12, 2019.

\bibitem[Zhang et~al.(2017)Zhang, Li, Wei, Chen, Qiu, Jiang, Yang, Zheng, Qin,
  Huang, et~al.]{zhang2017synergistic}
W.-J. Zhang, Y.~Li, M.-N. Wei, Y.~Chen, J.-G. Qiu, Q.-W. Jiang, Y.~Yang, D.-W.
  Zheng, W.-M. Qin, J.-R. Huang, et~al.
\newblock Synergistic antitumor activity of regorafenib and lapatinib in
  preclinical models of human colorectal cancer.
\newblock \emph{Cancer letters}, 386:\penalty0 100--109, 2017.

\bibitem[Zhao et~al.(2013)Zhao, Nishimura, Chen, Azeloglu, Gottesman,
  Giannarelli, Zafar, Benard, Badimon, Hajjar, et~al.]{zhao2013systems}
S.~Zhao, T.~Nishimura, Y.~Chen, E.~U. Azeloglu, O.~Gottesman, C.~Giannarelli,
  M.~U. Zafar, L.~Benard, J.~J. Badimon, R.~J. Hajjar, et~al.
\newblock Systems pharmacology of adverse event mitigation by drug
  combinations.
\newblock \emph{Science translational medicine}, 5\penalty0 (206):\penalty0
  206ra140--206ra140, 2013.

\bibitem[Zheng et~al.(2018)Zheng, Sun, and Simeonov]{zheng2018drug}
W.~Zheng, W.~Sun, and A.~Simeonov.
\newblock Drug repurposing screens and synergistic drug-combinations for
  infectious diseases.
\newblock \emph{British journal of pharmacology}, 175\penalty0 (2):\penalty0
  181--191, 2018.

\end{thebibliography}

%USE THE BELOW OPTIONS IN CASE YOU NEED NUMBERED FORMAT. UNCOMMENT THE ABOVE TWO LINES.
%\bibliographystyle{plain}
%\bibliography{reference}

%% sample for biography with author's image
\begin{biography}{}{\author{Jinxian Wang.} Jinxian Wang received the Bachelor's degree from hunan Agricultural University in 2019, and at present is studying for a master’s degree at Central South University supervised by Prof. Lei Deng. His study focuses on machine learning and bioinformatics.}
\end{biography}
\begin{biography}{}{\author{Xuejun Liu.} Xuejun Liu is a professor at School of Computer Science and Technology, Nanjing Tech University, Nanjing, China. His research interests include data mining and deep learning.}
\end{biography}
%% sample for biography without author's image
\begin{biography}{}{\author{Siyuan Shen.} Siyuan Shen is a graduate student at School of Software, Xinjiang University, Urumqi, China. His research interest is using machine learning algorithms to study non-coding RNA interactions and functions.}
\end{biography}
\begin{biography}{}{\author{Lei Deng.} Lei Deng is a professor at School of Computer Science and Engineering, Central South University, Changsha, China. His research interests include data mining, bioinformatics and systems biology.}
\end{biography}
\begin{biography}{}{\author{Hui Liu.} Hui Liu is a professor at School of Computer Science and Technology, Nanjing Tech University, Nanjing, China. His research interests include the anti-cancer drug screening by means of bioinformatics and deep learning.}
\end{biography}

\end{document}